\documentclass[10pt,twocolumn,letterpaper]{article}

\usepackage{cvpr}
\usepackage{times}
\usepackage{epsfig}
\usepackage{graphicx}
\usepackage{amsmath}
\usepackage{amssymb}

\usepackage[pagebackref=true,breaklinks=true,letterpaper=true,colorlinks,bookmarks=false]{hyperref}

\cvprfinalcopy 

\def\cvprPaperID{****} 

\ifcvprfinal\fi
\begin{document}

\title{VOC-ReID: Vehicle Re-identification based on Vehicle-Orientation-Camera}

\author{Xiangyu Zhu, Zhenbo Luo, Pei Fu, Xiang Ji \\
Ruiyan Technology \\
{\tt\small zhuxiangyu@ruiyanai.com}
{\tt\small luozhenbo@ruiyanai.com}
}

\maketitle

\begin{abstract}
Vehicle re-identification is a challenging task due to high intra-class variances and small inter-class variances. In this work, we focus on the failure cases caused by similar background and shape. They pose serve bias on similarity, making it easier to neglect fine-grained   information. To reduce the bias, we propose an approach named VOC-ReID, taking the triplet vehicle-orientation-camera as a whole and reforming background/shape similarity as camera/orientation re-identification. At first, we train models for vehicle, orientation and camera re-identification respectively. Then we  use orientation and camera similarity as penalty to get final similarity. Besides, we propose a high performance baseline boosted by bag of tricks and weakly supervised data augmentation. Our algorithm achieves the second place in vehicle re-identification at the NVIDIA AI City Challenge 2020. Codes are aviable at \url{https://github.com/Xiangyu-CAS/AICity2020-VOC-ReID}
\end{abstract}

\section{Introduction}
Vehicle re-identification aims to match vehicle appearances across multiple cameras. As the widely deployment of cameras throughout the cities’ roads, the task has more and more applications in smart cities, and also attracts more and more interest in the computer vision community [1, 2, 3]. 

Vehicle re-identification benefits from the progresses of person re-identification [4, 5, 6], network architectures [7, 8, 9], training loss [10, 11], data augmentation [12, 13]. But vehicle re-identification still has many challenges, including intra-class variations for vehicle-model, viewpoints, occlusion, motion blur [14].

From our observations, vehicle re-identification task suffers from the similar car shape and background. Images captured by camera share the same background, and orientation is a key-component of car shape. Thus, It is  reasonable to take orientation and camera into account as well as vehicle IDs. Based on the idea, we propose an mechanism named VOC-ReID to address the bias incurred by background and car shape. Our major contributions are threefold:

\begin{itemize}
\item We argue that the similarity of background can be interpreted as camera re-identification and similarity of shape interpreted as orientation re-identification. 
\item To reduce the bias of background and shape, we propose VOC-ReID approach which takes the triplet vehicle-orientation- camera as a whole.  To our best knowledge, we are the first to joint vehicle, orientation and camera information in vehicle re-identification task. 
\item Our approach achieves the second place in vehicle re-identification track at the NVIDIA AI City Challenge 2020 without using any extra data and annotations. It also out-performances the state-of-the-art methods on VeRi776 dataset [2].
\end{itemize}

\begin{figure*}[t]
\begin{center}
	\includegraphics[width=1\linewidth]{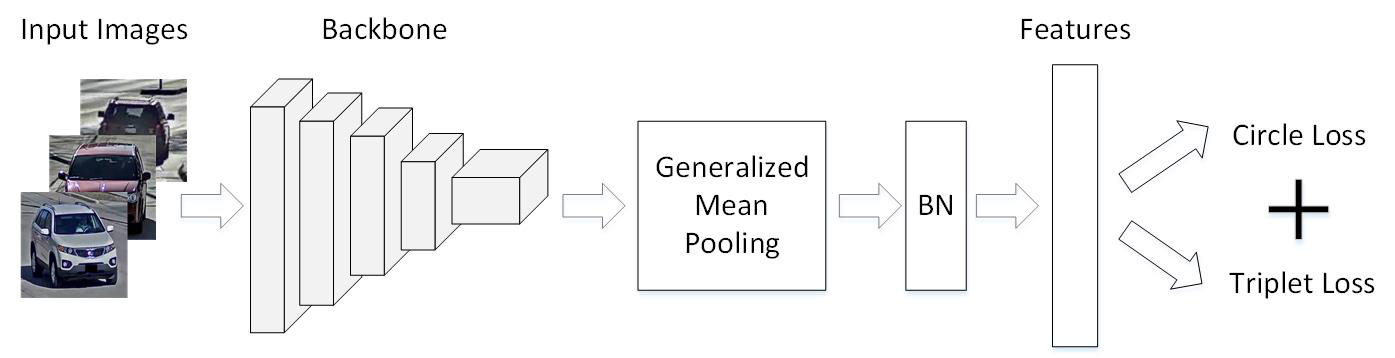}
\end{center}
\caption{the overall design of the training framework: RECT-Net, which stands for ResNet backbone, GeM pooling, Circle loss [8], and Triplet loss.}
\label{fig:architect}
\end{figure*}
\section{Related Works}
Recently, deep learning based vehicle Re-ID approaches outperform all previous baselines using handcrafted features. To learn a discriminative representation, a large-scale annotated data is needed. Liu et al. [2, 15] introduces a high-quality multi-view (VeRi776) dataset [2], which is the first large-scale benchmark for vehicle ReID.

An effective strategy to learn the discriminative representation is metric learning. Because the success of triplet embedding in person ReID field [10], Bai et al. [16] introduces Group-Sensitive triplet embedding to better model the intra-class variance. And Kumar et al. [14] provides an extensive evaluation of triplet loss applied to vehicle re-identification and study the sampling approaches for mining informative samples. 

On the other hand, some methods focus on exploiting viewpoint-invariant features, e.g. 2D key-points features. The approach by Tang et al. [18] that embeds key-points, heatmaps and segments from pose estimation into the multi-task learning pipeline for vehicle ReID, which guides the network to pay attention to viewpoint-related information. At the same time, [17, 18] use the graphic engine to generate the synthetic data with 2D key-points, colors and models, etc.. Similarly, Zhou et al. [19] uses a generative adversarial network to generate synthesize vehicle images with diverse orientation and appearance variations. 	However, these works did not take into account the bias of similar shapes and backgrounds.

\section{Proposed Approach}

In section 3.1, we introduce our baseline architecture using bag of tricks and data augmentation, which is a basic unit in section 3.2. Then the triplet re-identification system, abbreviated as VOC-ReID, adopting three units is described in section 3.2. In section 3.3, we introduce the training and post-processing tricks.

\subsection{Training Network}
Baseline architecture is shown in Figure \ref{fig:architect}. In the following paragraph, we use an abbreviation RECT-Net to represent it, which stands for the combination of ResNet [7, 8, 9], GeM pooling [23], Circle loss [11], and Triplet loss [10]. Data augmentation is also introduced in this section, including weakly supervised detection [12, 13], random erasing [24], color jittering and other methods.

{\bf ResNet backbone.} We use ResNet [7] with IBN [9] structure as feature extractor, which shows potentials in resisting viewpoint and illumination changes.

{\bf GeM: Generalized-mean Pooling.} Global average pooling is widely-used in classification tasks, but it is not a good choice for fine-grained instance retrieval tasks, just like vehicle re-id. In [23], generalized-mean (GeM) pooling is proposed as:

\begin{equation}
f^{(g)} = \left[f_1^{(g)}\dots f_k^{(g)}\dots f_K^{(g)}\right]^T, f_k^{(g)}=\left[\frac{1}{|X|_k }\sum_{x\in X}x^p\right]
\end{equation}

Average pooling and max pooling are all special cases of GeM pooling, it degrades to max pooling when $p\rightarrow \infty$ and degrades to average pooling when $p=1$. GeM pooling get superior performance than global average pooling in our vehicle re-id experiments.

{\bf Loss: Circle Loss and Triplet Loss.} We train our networks with two types of losses: Circle loss [11] and Triplet loss [10]. The Circle loss has a unified formula for two elemental deep feature learning approaches, i.e., learning with class-level labels and pair-wise labels [11]. The hyper-parameters of Circle loss are decided after several experiments. The scale factor is 64 and the relaxation factor is 0.35. Triplet loss is widely used in re-id tasks. In [10], the batch-hard triplet loss selects only the most difficult positive and negative samples. The hyper-parameter of batch-hard triplet loss, the margin, is 0.3. At last, we train our networks with the fusion of Circle loss and Triplet loss, with the coefficients of 1:1.

{\bf Data Augmentation: Weakly Supervised Detection.} Data augmentation is widely used to improve the models’ accuracy and avoid overfitting. One important data augmentation works in our approach is weakly supervised detection augmentation. Inspired by [12, 13], we train an initial vehicle ReID model to get heatmap response of each image and setting a threshold to get the bounding box larger than it. Compared with supervised detection methods, it produces a more tightly cropping of vehicle, focusing on the attentional areas of Re-ID model. After weakly supervised detection, we get a cropped copy of both train and test set. Train set is used alongside with original images. So after applying weakly supervised data augmentation, our dataset get doubled.
Other commonly used data augmentation methods are applied in our training stage, such as random erasing [24], and color jitter, etc.

\subsection{VOC-ReID}
Similar background and shape pose sever bias on final similarity matrix. Thus, similarity contributed by fine-grained information would be dominated by background and shape similarity. To reduce the bias so that the model can focus on fine-grained details, we propose the triplet vehicle-orientation-camera re-identification, abbreviated as VOC-ReID.

\begin{figure}[t]
\begin{center}
	\includegraphics[width=1\linewidth]{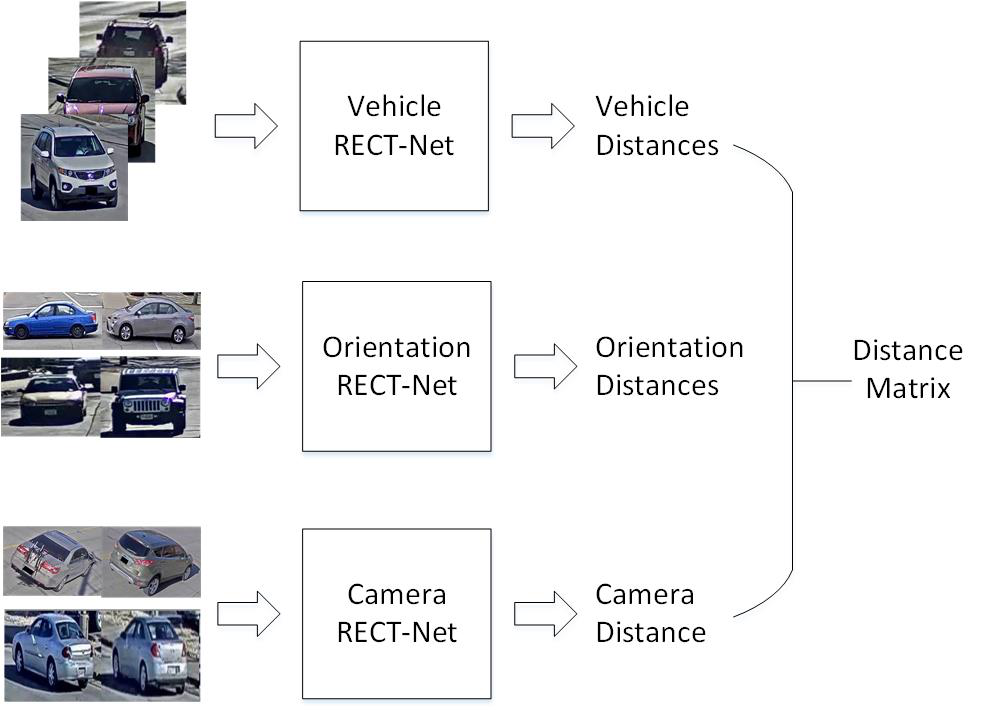}
\end{center}
\caption{the overall framework of the proposed VOC-ReID system. It includes three parts: Vehicle RECT-Net, output the vehicle similarity; Orientation RECT-Net, output the orientation similarity; Camera RECT-Net, output the camera similarity. At last, we use a straight-forward distance fusion method to fuse these 3 outputs together to  get the last output distance matrix.}
\label{fig:voc}
\end{figure}

As illustrated in Figure \ref{fig:voc}, the VOC-ReID system extracts the vehicle feature distance matrix, orientation feature, distance matrix, camera feature distance matrix separately, and then fuses them to one distance matrix. Compared with learning vehicle IDs alone, vehicle-orientation-camera triplet can boost the performance significantly. 

\begin{figure}[t]
\begin{center}
	\includegraphics[width=1\linewidth]{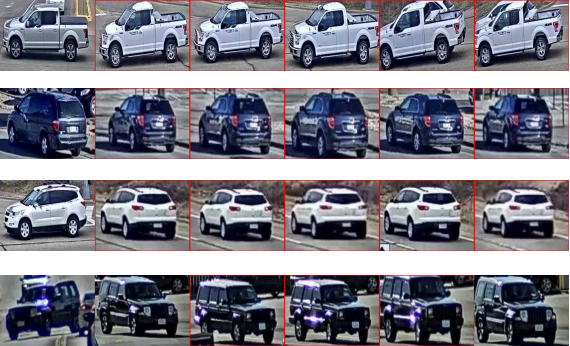}
\end{center}
\caption{Some false predicts by Vehicle RECT-Net (Red rectangle means the false predicts). As row a and b show: Vehicle RECT-Net usually predicts the vehicle with the similar orientation as the same vehicle ID. As row c and d show: Vehicle RECT-Net usually predicts the vehicle with the similar background as the same vehicle ID. }
\label{fig:failure}
\end{figure}

{\bf Vehicle ReID.} In vehicle ReID part, vehicle ID is used as the ground-truth to optimize the 
distances between vehicles. The purpose of Vehicle ReID is to get a compact embedding $f_{v}(x_i)$. Cosine distance  between image $x_i$ and $x_j$  can be computed by the following equation:
\begin{equation}
{D_v (x_i, x_j)}=\frac{f_v(x_i) \cdot f_v(x_j)}{|f_v(x_i)| \cdot |f_v(x_j)|}
\end{equation}
Although our baseline architecture RECT-Net show superior performance on both person and vehicle Re-ID datasets, it still can’t handle some typical failure cases, shown in Figure \ref{fig:failure}. Those failure cases all have similar orientation or background.

That phenomenon encourages us to design another two networks to learn the orientation similarity and background similarity.

{\bf Orientation ReID.} In this section, we reform the learning of shape similarity as an orientation Re-ID problem. It’s hard to define shape similarity directly. A precise sign of shape is vehicle key-points [25], but it’s confusing to define pose similarity. So, finally we choose orientation as a rough but stable representation of shape. After replacing vehicle IDs to orientation ID, we can practice an orientation ReID process, similar to orientation estimation. However, rather than outputting a regression result or a classified direction, orientation ReID outputs an embedding that can be used to calculate orientation similarity. In this work, we use a completely same architecture RECT-Net to learn orientation similarity. 

To train the Orientation RECT-Net, we need to get data with orientation labels. Fortunately, synthetic data VehicleX [8] has orientation label available, avoiding manually labeling work. In more detail, we quantize the angle (0 - 360) into 36 bins, each bin is treated as an ID. After preparing the training data set, we can train the Orientation RECT-Net on VehicleX dataset and calculate orientation similarity on CityFlow. There is domain gap between CityFlow and VehicleX, so the predicted orientation accuracy can’t be guaranteed. However, we don’t need a precise prediction of orientation, a rough orientation similarity is qualified to reduce similarity bias caused by shape.

We get the orientation feature $f_o(x_i)$ from RECT-Net, and then orientation distance  between the vehicle image  $x_i$ and $x_j$ can be computed: 
\begin{equation}
D_o (x_i, x_j)=\frac{f_o(x_i) \cdot f_o(x_j)}{|f_o(x_i)| \cdot |f_o(x_j)|}
\end{equation}
{\bf Camera ReID.} In this section we reform the learning of background similarity as a camera ReID problem. According to prior knowledge, images captured by the same camera share similar background, as well as image style and illumination condition. Thus, camera ID is a reliable label to indicate background variances. Nearly every Re-ID dataset provide a camera ID as well as vehicle or person ID, but few works have utilized it properly.

Let’s define two vehicle images as $(x_i, x_j), x_i\in (c_a), x_j\in (c_b), a, b \in[0, N]$, $c_a$ means the camera $a$, then the similarity of background can be roughly represented by camera.  As the training set already has the camera ID label, we can simply use the same network architecture of Vehicle RECT-Net to train the Camera ReID. Then we can extract the camera-aware features $f_c(x_i)$, and the camera-aware 
distance between the vehicle image  and  can be computed: 
\begin{equation}
D_c (x_i, x_j)=\frac{f_c(x_i) \cdot f_c(x_j)}{|f_c(x_i)| \cdot |f_c(x_j)|}
\end{equation}

Some previous work also try to explore the usage of camera ID, such as He et al. [26] use the camera ID information as a prior information, they assume that if two vehicle images are from the same camera, they should not be the same vehicle, that means:  

\begin{equation}
D(x_i, x_j)=
\begin{cases}
0, \quad &c_a =c_b\\
1, \quad &c_a \neq c_b\\
\end{cases}
\end{equation}
Such rule is a hard discrimination, it may cause other problem and the threshold varies among datasets, which means it should be carefully adjusted.

Our approach considers camera information as a complement to the overall information, and it is more reasonable. As far as we know, it is the first time to utilize the camera ID information in training and infer procedure. 

{\bf VOC Fusion.}  As the figure 4 shows, in the test phase, according to the equations (1) (2) (3), the VOC-ReID system will output the vehicle ID distance matrix, orientation distance matrix, camera distance matrix, we need to fuse these three distance matrix to be one distance matrix. The fusion distance matrix can be expressed as following:

\begin{equation}
D(x_i, x_j) = D_v(x_i, x_j) - \lambda _1 D_o(x_i, x_j) - \lambda _2 D_c(x_i, x_j)
\end{equation}
As analyzed in the section 3.2.1, it should decrease the influence of the similar orientation and background, and the three distances matrix are all cosine distance, so we can directly minus the orientation distance and camera distance. In experiment, we set $\lambda _1 = \lambda _2 = 0.1$.

\subsection{Training and Testing Tricks}

{\bf Cosine learning rate decay. } In training phase, we adopt cosine annealing scheduler to  decay learning rate. According to the experiments conducted by He et al. [27], multi-step scheduler inferior performance compared with cosine annealing. Cosine decay starts to decay the learning since the beginning but remains large until step decay reduces the learning rate by 10x, which can potentially improve the training progress. 

{\bf Image to track.} In the test phase, the traditional solution is to compute the distance between query images and gallery images. As the images in one track are the same vehicle, and one track covers multiple similar images but may have different perspective, so we can replace the features of each gallery image with the average features of the track, which can be regarded as an improvement from image-to-image to image-to-track way.

{\bf Re-rank.} When get the fusion distance matrix, then we adopt the re-ranking method [28] to update the final result, which will significantly improve the mAP.

\section{Experiments}
\subsection{Implementation Details}
All input images are resized to 320x320 and then applied with a series of data augmentation methods in training, including random horizontal flip, random crop, color jitter and random erase. We also try other data augmentation methods like random patch, augmix [29], random rotate/scale and random blur, but all of them turn out not working well on this task.
To meet the requirement of pair-wise loss, data is sampled by m-per-class sampler with parameter P and K, which denote numbers of classes and numbers of instances per class in a mini-batch. In our experiments they are set to P=4, K=16. In train process, ResNet50-IBN-a is used as backbone for both vehicle and orientation/camera ReID. All models are trained on a single GTX-2080-Ti GPU in total 12 epochs, with feature layers frozen in the first epoch, which works as a warm-up strategy. Cosine annealing scheduler is adopted to decay initial learning rate from 3.5e-4 to7.7e-7.

\subsection{Dataset}
There are totally 666 IDs and 56277 images in CityFlow. 36935 images with 333 IDs are used for training and the others for testing. A new rule in AI City Challenge 2020 that prohibits using extra data, such as VeRi776 [2] and VehicleID[30].
To overcome the shortness of lacking data, the official committees provide a synthetic vehicle ReID dataset named VehicleX [17]. It contains 1362 IDs and 192150 images generated by Unity engine, with high orientation and light variance. We only use the trainset of CityFlow and the whole synthetic dataset VehicleX.

\subsection{Ablation Study}
In this section, we split the trainset of CityFlow into two parts, the first 95 IDs are used as validation set while the rest for training.

{\bf Influences of Losses.} Cross Entropy loss combined with Triplet loss are widely used as baseline in person and vehicle re-identification. Recent years tremendous loss functions emerge in face recognition and metric learning domains, lots of them stand the test of time and finally prove to be quite solid. In our experiments, We compare Cross Entropy (CE) with other popular classification learning losses, such as Arcface, Cosface and Circle loss, in Table \ref{tab:loss}. It shows that Circle loss alongside with Triplet loss achieves the best result with large margin just as the author claims. 

\begin{table}
\begin{center}
\begin{tabular}{|c|cc|}
\hline
loss & mAP & Rank1\\
\hline\hline
CrossEntropy + Triplet & 21.6\% & 42.1\% \\
Arcface + Triplet & 26.2\% &46.7\%\\
Cosface + Triplet & 26.8\% &48.3\%\\
Circle + Triplet & 29.7\% &50.8\%\\
\hline
\end{tabular}
\end{center}
\caption{Different losses on CityFlow validation set.}
\label{tab:loss} 
\end{table}

{\bf Influences of Data.} PAMTRI [18] and VehicleX have fully explored the proper usage of synthetic data in Vehicle ReID. In Table \ref{tab:data}., Real denotes real dataset namely trainminusval set of CityFlow,  VehicleX denotes synthetic dataset VehicleX. Aug denotes weakly supervised detection augmentation described in section 3.1. As we can see, by adding synthetic dataset VehicleX, our method achieves a much better performance. On the other hand, weakly supervised data augmentation brings almost 5\% gains because of relatively loose cropping in CityFlow. The scale factor of vehicle and background would obviously deteriorate the performance.

\begin{table}
\begin{center}
\begin{tabular}{|c|cc|}
\hline
Data & mAP & Rank1\\
\hline\hline
Real & 29.7\% & 50.8\% \\
Real + VehicleX & 39.5\% &64.0\%\\
Real + Aug + VehicleX & 44.4\% &65.3\%\\
\hline
\end{tabular}
\end{center}
\caption{Different losses on CityFlow validation set.}
\label{tab:data}
\end{table}

{\bf Influences of VOC fusion.} To explore the effectiveness of VOC-ReID mechanism, we apply orientation and camera similarity penalty respectively. In Table \ref{tab:voc}. RSA denotes Real data, Synthetic data and weakly supervised augmentation, orientation denotes orientation aware ReID and camera denotes camera aware ReID. After applying orientation and camera similarity penalty, rank1 increases 10\%, from 65.3\% to 75.5\%.

\begin{table}
\begin{center}
\begin{tabular}{|l|c c c| }
\hline
Method & \multicolumn{3}{|c|}{Performance} \\
\hline\hline
Vehilce ReID &  $\checkmark$ & $\checkmark$ &$\checkmark$ \\
Orientation ReID & &$\checkmark$ &$\checkmark$\\
Camera ReID  & & &$\checkmark$\\

\hline
mAP & 44.4\% & 47.0\% & 50.8\%\\
rank1 &65.3\% &70.5\% &75.5\% \\

\hline
\end{tabular}
\end{center}
\caption{Influences of VOC fusion on CityFlow validation set.}
\label{tab:voc}
\end{table}

\subsection{Performance on AICity 2020 Challenge}

\begin{figure}[htb]
\begin{center}
	\includegraphics[width=1\linewidth]{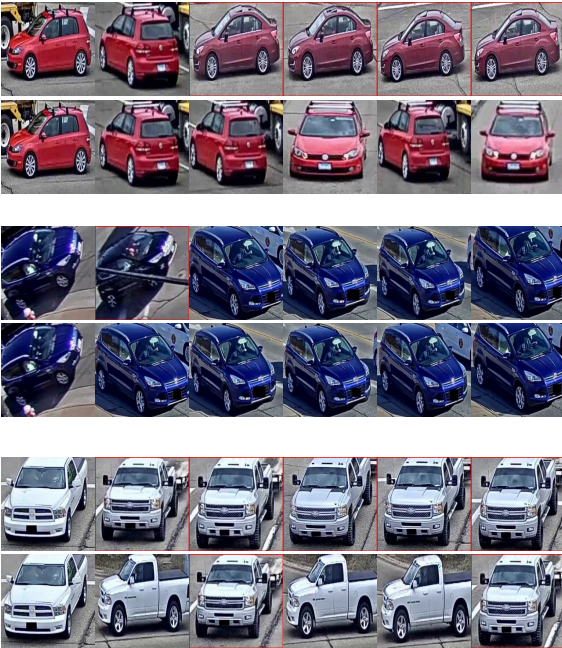}
\end{center}
\caption{Failure cases that rectified by VOC-ReID. Each group indicates a comparison between before and after applying orientation/camera aware ReID. The first row is the result without orientation/camera ReID while the second row indicates the result with orientation and camera ReID}
\label{fig:rectify}
\end{figure}

In the CVPR 2020 AI City Challenge Image-based Vehicle ReID Track (Track2), our submission ranks the second place on final leaderboard, without using any extra data or annotations. Performance of top 10 teams are shown in Table \ref{tab:team}.

Then final result ensembles five models, there are three models for vehicle ReID, one for orientation aware ReID and one for camera aware ReID. The three vehicle ReID models are trained on the same dataset with different backbones, including resnet50-ibn-a, resnet101-ibn-a and resnext101-ibn-a. Simply ensemble models with different backbones does not benefit a lot, it approximately improves 2\% on mAP at the cost of tripling training processes and model sizes, which totally not worth it. 

Figure . shows some failure cases that rectified by applying orientation and camera aware ReID. Observing from Figure 4, we can see a lot of failure cases have the same orientations and captured from the same cameras, which contribute to the final similarity. 

\begin{table}
\begin{center}
\begin{tabular}{|l|c c|}
\hline
Rank  & Team Name & mAP \\
\hline\hline
1  & Baidu-UTS &84.1\% \\
2 &\textbf{RuiYanAI(Ours)} &\textbf{78.1}\% \\
3  & DMT &73.1\%\\
4  & IOSB-VeRi &68.9\%\\
5  & BestImage &66.8\%\\
6  & BeBetter &66.6\%\\
7  & UMD &66.6\%\\
8  & Ainnovation &65.6\%\\
9  & NMB &62.0\%\\
10 & Shahe &61.9\%\\

\hline
\end{tabular}
\end{center}
\caption{Leaderboard of AI City 2020 Track2}
\label{tab:team}
\end{table}

\subsection{Performance on VeRi776}
We also test our method on VeRi776 dataset and compare with state-of-the art methods in Table \ref{veri}. Our baseline model outperformances other methods, and applying orientation/camera aware penalty can further boost the performance. We notice that the hyper-parameter P and K in M-Per-Classs sampler largely affect results, on the other hand larger resolution benefits results.

\begin{table}
\begin{center}
\begin{tabular}{|l|c c|}
\hline
Method  & mAP & Rank1 \\
\hline\hline
StrongBaseline [4] &67.6\% &90.2\% \\
DMML [5] &70.1\% &90.2\% \\
PAMTRI(ALL) [3] &71.8\% &92.8\% \\
RECT-Net(P=4, K=16, size=256) &78.6\% &95.9\% \\
RECT-Net(P=4, K=16, size=320) &81.6\% &96.8\%\\
+Orientation ReID &82.8\% &97.6\% \\

\hline
\end{tabular}
\end{center}
\caption{Comparison with state-of-the art methods on VeRi776}
\label{tab:veri}
\end{table}

\section{Conclusion}
In this paper, we propose a vehicle ReID approach VOC-ReID, which is a framework that joint the vehicle, orientation and camera information together. Previous works either focus on one aspect or spatial-temporary constraints, or joint the view-point invariant feature (2D/3D keypoints, type, color) with the vehicle information to match the vehicle identities. However, we notice that the similarity of vehicle shape and background are key factors that affecting the final result. Therefore, We argue that the similarity of background can be interpreted as camera re-identification and similarity of shape interpreted as orientation re-identification, then we propose VOC-ReID approach which takes the triplet vehicle-orientation- camera as a whole to reduce the bias caused by the similar vehicle shape and background.. Finally, extensive experiments are conducted on CityFlow and VeRi776 to evaluate the VOC-ReID mechanism against the-state-of-the-art methods. Our proposed framework achieves the top performance in VeRi776, and achieve 2nd place in AI City 2020 Track 2, and ablation studies show that each proposed component helps to enhance performance. 

We believe that the proposed method can generalized to  person re-identification task. In the future works, we are going to ensemble the triplet vehicle-orientation-camera in a unified network and explore its effectiveness in person re-identification.

\subsection*{Reference}

\noindent[1] H. Liu, Y. Tian, Y. Wang, L. Pang, and T. Huang. Deep relative distance learning: Tell the difference between similar vehicles. In Proceedings of the IEEE Conference on Computer Vision and Pattern Recognition, pages 2167-2175, 2016.

\noindent[2] X. Liu, W. Liu, H. Ma, and H. Fu. Large-scale vehicle reidentification in urban surveillance videos. In Multimedia and Expo (ICME), 2016 IEEE International Conference on, pages 1–6, 2016.

\noindent[3] Z. Tang, M. Naphade, M.-Y. Liu, X. Yang, S. Birchfield, S. Wang, R. Kumar, D. C. Anastasiu, and J.-N. Hwang.  Cityflow: A city-scale benchmark for multi-target multicamera vehicle tracking and re-identification. In Proceedings of the IEEE Conference on Computer Vision and Pattern Recognition, 2019.

\noindent[4]Y. Sun, L. Zheng, Y. Yang, Q. Tian, and S. Wang. Beyond part models: Person retrieval with refined part pooling. In Proceedings of the European Conference on Computer Vision, pages 480–496, 2018.

\noindent[5]L. Zheng, H. Zhang, S. Sun, M. Chandraker, Y. Yang, and Q. Tian. Person re-identification in the wild. In Proceedings of the IEEE Conference on Computer Vision and Pattern Recognition, pages 1367-1376, 2017.

\noindent[6]Z. Zheng, X. Yang, Z. Yu, L. Zheng, Y. Yang, and J. Kautz. Joint discriminative and generative learning for person re-identification. In Proceedings of the IEEE Conference on Computer Vision and Pattern Recognition, pages 2138-2147, 2019.

\noindent[7]K. He, X. Zhang, S. Ren and J. Sun. Deep residual leanring for image recognition. In Proceedings of the IEEE Conference on Computer Vision and Pattern Recognition, pages770-778, 2016. 

\noindent[8]S. Xie, R, Girshick, P. Dollar, Z. Tu, and K. He. Aggregated Residual Transformations for Deep Neural Networks. arXiv 1161.05431, 2016.

\noindent[9]X. Pan, P. Luo, J. Shi, X. Tang. Two at Once: Enhancing Learning and Generation Capacities via IBN-Net. In Proceedings of the European Conference on Computer Vision, 2018.

\noindent[10]A. Hermans, L. Beyer, and B. Leibe. In defense of the triplet loss for person re-identification. arXiv:1703.07737, 2017.

\noindent[11]Y. Sun, C. Cheng, Y. Zhang, C. Zhang, L. Zheng, Z. Wang, Y. Wei. Circle Loss: A Unified Perspective of Pair Similarity Optimization. In Proceedings of the IEEE Conference on Computer Vision and Pattern Recognition, 2020

\noindent[12]T. Hu, H. Qi, Q. Huang, Y. Lu. See Better Before Looking Closer: Weakly Supervised Data Augmentation Network for Fine-Grained Visual Classification. arXiv 1901.09891v2, 2019.
\noindent[13]Y. Zhu, Y. Zhou, Q. Ye, Q. Qiu, and J. Jiao. Soft Proposal Networks for Weakly Supervised Object Localization. In Proceedings of the IEEE International Conference on Computer Vision, 2017

\noindent[14]R. Kumar, E. Weill, F. Aghdasi, P. Sriram. A Strong and efficient baseline for vehicle re-identification using deep triplet embedding. Journal of Artificial Intelligence and Soft Computing Research, Pages 27-45, 2020.

\noindent[15]X. Liu, W. Liu, T. Mei, and H. Ma. PROVID: Progressive and multimodal vehicle reidentification for large-scale urban surveillance. TMM, 20(3):645–658, 2017.

\noindent[16]Y. Bai, Y. Lou, F. Gao, S. Wang, Y. Wu, and L.Y. Duan. Group sensitive triplet embedding for vehicle re-identification. TMM, 20(9):2385–2399, 2018.

\noindent[17]Y. Yao, L. Zheng, X. Yang,  M. Naphade, and T. Gedeon. Simulating Content Consistent Vehicle Datasets with Attribute Descent. arXiv preprint arXiv:1912.08855. 2019

\noindent[18]T. Zheng, M. Naphade, S. Birchfifield, J. Tremblay, W. Hodge, R. Kumar, S. Wang and X. Yang. Pamtri: Pose-aware multi-task learning for vehicle re-identification using highly randomized synthetic data. In Proceedings of the IEEE International Conference on Computer Vision:211--220. 2019.

\noindent[19]Y. Zhou and L. Shao. Aware attentive multi-view inference for vehicle re-identification. In Proceedings of the IEEE Conference on Computer Vision and Pattern Recognition, pages 6489–6498, 2018.

\noindent[20]H. Luo, Y. Gu, X. Liao, S. Lai, W. Jiang. Bag of Tricks and A Strong Baseline for Deep Person Re-identification. In Proceedings of the IEEE Conference on Computer Vision and Pattern Recognition Workshops, 2019.

\noindent[21]H. Lawen, Avi Ben-Cohen, M. Protter, I. Friedman, Lihi Zelnik-Manor. Attention Network Robustification for Person ReID. arXiv 1910.07038v2, 2019.

\noindent[22]M. Ye, J. Shen, G. Jin, T. Xiang, L. Shao and S. C.H.Hoi. Deep Learning for Person Re-identification: A Survey and Outlook. arXiv 2001.04193v1, 2020.

\noindent[23]F. Radenovic, G. Tolias, and O. Chum. Fine-tuning cnn image retrieval with no human annotation. IEEE Transaction on Pattern Analysis and Machine Intelligence, vol. 41, no. 7, pages1655-1668, 2018.

\noindent[24]Z. Zhong, L. Zheng, G. Kang, S. Li and Y. Yang. Random erasing data augmentation. arXiv 1708.04896, 2017.

\noindent[25]Z. Wang, L. Tang, X. Liu, Z. Yao, S. Yi, J. Shao, J. Yan, S. Wang, H. Li, and X. Wang. Orientation invariant feature embedding and spatial temporal regularization for vehicle re-identification. In Proceedings of the IEEE International Conference on Computer Vision, pages 379–387, 2017.

\noindent[26]Z. He, Y. Lei, S. Bai, W. Wu. Multi-Camera Vehicle Tracking with Powerful Visual Features and Spatial-Temporal Cue. In IEEE Conference on Computer Vision and Pattern Recognition (CVPR) Workshops, June, 2019

\noindent[27]T. He, Z. Zhang, H. Zhang, Z.Zhang, J. Xie, M. Li. Bag of Tricks for Image Classification with Convolutional Neural Networks. arXiv preprint arXiv:1812.01187. 2019

\noindent[28]Z. Zhong, L. Zheng, D. Cao, and S. Li. Reranking person re-identification with k-reciprocal encoding. In Proceedings of the IEEE Conference on Computer Vision and Pattern Recognition, pages 1318–1327, 2017.

\noindent[29]H. Dan, M. Norman, C. D., Z. Barret, G. Justin and L. Balaji. AugMix: A Simple Data Processing Method to Improve Robustness and Uncertainty. In Proceedings of the International Conference on Learning Representations (ICLR), 2020.

\noindent[30]H. Liu, Y. Tian,  Y. Wang, L. Pang, and T. Huang. Deep Relative Distance Learning: Tell the Difference Between Similar Vehicles. In Proceedings of the IEEE Conference on Computer Vision and Pattern Recognition, pages 2167--2175, 2016.

\end{document}